%% file: main.tex
\documentclass[
]{ceurart}

\usepackage{listings}
\begin{document}

\copyrightyear{2024}
\copyrightclause{Copyright for this paper by its authors.
  Use permitted under Creative Commons License Attribution 4.0
  International (CC BY 4.0).}

\conference{xAI'24: 2nd World Conference on eXplainable Artificial Intelligence,
  July 17--19, 2024, Valletta, Malta\\\textbf{Accepted Manuscript}}

\title{Towards Assurance of LLM Adversarial Robustness using Ontology-Driven Argumentation}


\author[1]{Tomas Bueno Momcilovic}[%
]
\cormark[1]
\address[1]{fortiss GmbH Research Institute, Munich, Germany}

\author[2]{Beat Buesser}[%
]
\address[2]{IBM Research Europe, Zurich, Switzerland}

\author[3]{Giulio Zizzo}[%
]
\address[3]{IBM Research Europe, Dublin, Ireland}

\author[3]{Mark Purcell}[%
]

\author[1]{Dian Balta}[%
]

\cortext[1]{Corresponding author.}

\begin{abstract}
  Despite the impressive adaptability of large language models (LLMs), challenges remain in ensuring their security, transparency, and interpretability. Given their susceptibility to adversarial attacks, LLMs need to be defended with an evolving combination of adversarial training and guardrails. However, managing the implicit and heterogeneous knowledge for continuously assuring robustness is difficult. We introduce a novel approach for assurance of the adversarial robustness of LLMs based on formal argumentation. Using ontologies for formalization, we structure state-of-the-art attacks and defenses, facilitating the creation of a human-readable assurance case, and a machine-readable representation. We demonstrate its application with examples in English language and code translation tasks, and provide implications for theory and practice, by targeting engineers, data scientists, users, and auditors.
\end{abstract}


\maketitle

\section{Introduction}
\input{content_short/introduction}

\section{Background and Related Work}
\input{content_short/background}

\section{Assurance with Ontology-Driven Arguments}
\input{content_short/results}

\section{Conclusion}
\input{content_short/conclusion}

\begin{acknowledgments}
 This work was partially supported by financial and other means by the following research projects: DUCA (EU grant agreement 101086308), and FLA (supported by the Bavarian Ministry of Economic Affairs, Regional Development and Energy). We thank the reviewers for their valuable comments. 
\end{acknowledgments}

\bibliography{refs}

\end{document}

%% file: content_short/introduction.tex
Large language models (LLMs) have shown promise in various natural and domain-specific language tasks \cite{fan2023,shi2022}, even without further training \cite{kojima2022}. However, challenges hinder their trustworthiness \cite{eg2019liability}, as LLMs have an inscrutable structure and dynamicity that make them a moving target for safety and security research \cite{bommasani2023}. In particular, they are brittle against adversarial attacks; slight perturbations in the input can cause a model to provide malicious output \cite{Zou2023_Universal}, and guardrails can often only be introduced post-incident \cite{robey2023smoothllm}.

Given the novelty and fast-paced evolution of LLMs, engineers need to rely on preprints and experiments (cf. \cite{geiping2024}) to analyse the impact of novel attacks and envision suitable defenses. Unlike software security, for which maintained knowledge bases exist (e.g. Common Vulnerability Enumerations \cite{mitre2023}), no such process is established for LLMs. Consequently, the required knowledge is captured in the data, code, documentation, and brains of individuals. This implicit knowledge base for assurance may not capture the entire picture of attacks and confidence in defenses over time. For instance, a very recent example by Microsoft shows extremely effective ``multi-turn jailbreaks" across LLMs, which would require engineers to redesign the existing defenses by combining heterogeneous knowledge: attack model and history analysis, prompt and response analysis, turn-pattern analysis, turn-by-turn and overall defenses \cite{russinovich2024great}.

Hence, the question we seek to address is: \textit{How can one continuously assure that an LLM is robust enough against adversarial attacks in a particular domain?} In this research-in-progress work, we propose an assurance approach that allows for structuring the heterogeneous knowledge about LLM attacks and defenses (cf. \cite{silva2022}), as well as the application domain. We handle the knowledge involved in creating assurance arguments explicitly and comprehensively based on ontological models. The latter allow for a formal argumentation along human-readable assurance cases expressed with machine-readable ontologies, thus creating a shared understanding about training, guardrails, and implementation.

%% file: content_short/background.tex
\textbf{LLMs} are neural network models that are pre-trained on a large amount of text data and have been shown to be capable of predicting, translating, or generating text for natural \cite{shi2022} and programming languages \cite{fan2023}. 

Traditional \textbf{adversarial attacks} add imperceptible perturbations $\delta$ to a given data point $x$ so that a classifier $f$ predicts $f(x) = c$ and $f(x+\delta) = c'$ where $c \neq c'$. Attacks on LLMs involve malicious prompts bypassing guardrails or model alignment to obtain harmful outputs \cite{Zou2023_Universal}. Obtaining such prompts includes gradient-based optimizations of the input \cite{Zou2023_Universal}, persuasion patterns to bypass guardrails \cite{Zeng2024_How}, and model inversions to generate vulnerable code in non-natural-language tasks \cite{hajipour2023systematically}. \textbf{Robustness defenses} are similarly developing and highly heterogeneous; they include, for example, perplexity filters against gradient-based suffix-style attacks \cite{alon2023detecting}, estimation of the brittleness of jailbreaks \cite{robey2023smoothllm}, and instructions for LLMs to detect harmful prompts \cite{helbling2023llm}.

\textbf{Assurance} is the process of structuring an argument from claims about a system and its environment that are grounded by evidence \cite{acwg2021}. An \textbf{assurance case} is a bundle of arguments, used to assess the level of confidence in a particular quality of a system \cite{Batarseh2021} in a domain. Assurance cases have been shown to be suitable for complex and rapidly evolving AI technologies \cite{hawkins2021}, and also usable for structuring claims about explainability and interpretability \cite{silva2022}.

Assurance of AI security draws on traditional methods such as verification with test libraries \cite{Nicolae2019}, validation with human feedback \cite{MacGlashan2017}, and manual \cite{wei2024} and automated \cite{madry2017towards} stress testing. However, the inscrutability of AI has motivated the proliferation of experimental interpretability \cite{Rauker2023}, auditing \cite{koshiyama2021towards}, and forensic \cite{shan2022} methods to investigate the causes of problematic output. Research which makes use of both approaches includes the work of Kläs \textit{et al.} \cite{klas2021} on risk-based assurance cases for autonomous vehicles, and Hawkins \textit{et al.} \cite{hawkins2021} on a dynamic assurance framework for autonomous systems.

Since arguments may cover heterogeneous knowledge about the technology and its domain, \textbf{knowledge formalization} proves valuable for creating a common understanding. Knowledge representation and reasoning is a field of AI research \cite{delgrande2023current}, covering topics such as formalization based on ontologies to support explainable AI \cite{chari2020explanation}. An \textbf{ontology} is “an explicit specification of a conceptualization” (p. 199, \cite{gruber1993}) that allows machine-readable knowledge to be shared between humans in a common vocabulary. 

While the combination of ontologies and assurance cases is not entirely novel - Gallina \textit{et al.} \cite{gallina2023} propose such a framework for assuring AI conformance with the EU Machinery Regulation - we note that continuously formalizing, assuring and reasoning about LLM security is a novel proposition. Our approach links two graph representations in the same ontology: a non-hierarchical, mixed-direction acyclic graph of attacks and defenses in the LLM's application domain, and a hierarchical directed acyclic graph of corresponding claims and evidence about its robustness. The elements of both graphs are represented as subject-predicate-object semantic triples using the Resource Description Framework \cite{w3_2014_rdf} and Web Ontology Language \cite{w3_2012_owl}. We additionally make use of the Goal Structuring Notation (GSN) metamodel \cite{acwg2021} to structure assurance cases (cf. Figure \ref{fig2}) with goals (G), strategies (S), solutions (Sn), contexts (C) and justifications (J), and add attacks as counterclaims (CC) following community practice \cite{bloomfield2022}.

%% file: content_short/results.tex
\subsection{Robustness in Natural Language Tasks}

Recent experiments show that simple attacks can have high success rates in the natural language application domain \cite{Zou2023_Universal}. For example, Geiping \textit{et al.} \cite{geiping2024} demonstrate that in most tests, particular characters in seemingly benign prompts (e.g., Latin, Chinese, ASCII) can successfully induce a particular response from many pre-trained open-source LLMs (e.g., LLaMa-2 with 7 billion parameters). For example, an attack is deemed successful if an LLM responds with profanities (i.e., profanity attacks) or reveals its hidden system instruction (i.e., extraction attacks).

Several options can help reduce the vulnerability of an LLM to such attacks. Retraining the LLM to be robust to character-specific perturbations \cite{cao2023defending} is arguably more secure than simply filtering the input based on prompt properties \cite{alon2023detecting}, but also more resource and time intensive. Thus, an engineer may decide to combine defenses in stages: add a naive input filter to exclude prompts with reportedly ``risky" character types in the short-term; perform experiments with benign and adversarial prompts, reconfiguring the filter to adjust the parameters according to results in the medium-term; and adversarially retrain the LLM to be deployed in the longer-term.

\begin{figure}[]
\centering
\includegraphics[width=\textwidth]{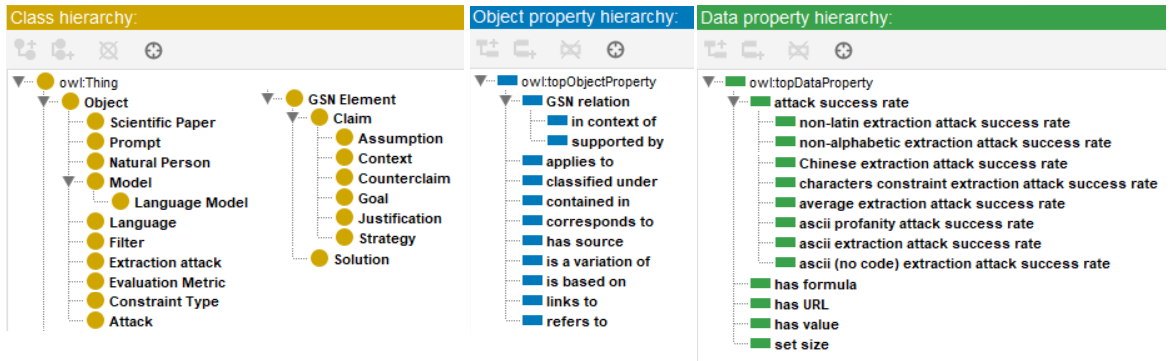}
\caption{Structure of the ontology. Visualized using Stanford Protégé.} \label{fig3}
\end{figure}

We develop an ontology that formalizes the relations between concepts (i.e., LLM, attack and constraint type) and variables (i.e., attack success rate, character) as described in the paper \cite{geiping2024}. The ontology is implemented with a trivial structure (cf. Figure \ref{fig3}), consisting of classes, object properties (i.e., relations) and data properties (i.e., values). In the example provides the attack success rate (e.g., \texttt{String1\_ASR: 0.5}) of an individual attack (e.g., adversarial extraction-type prompt with Chinese-English characters) with the LLM (e.g., LLaMa-2-7B-chat) and the constraint under which the attack functions (e.g., Chinese language characters).

\begin{figure}[]
\centering
\includegraphics[width=0.80\textwidth]{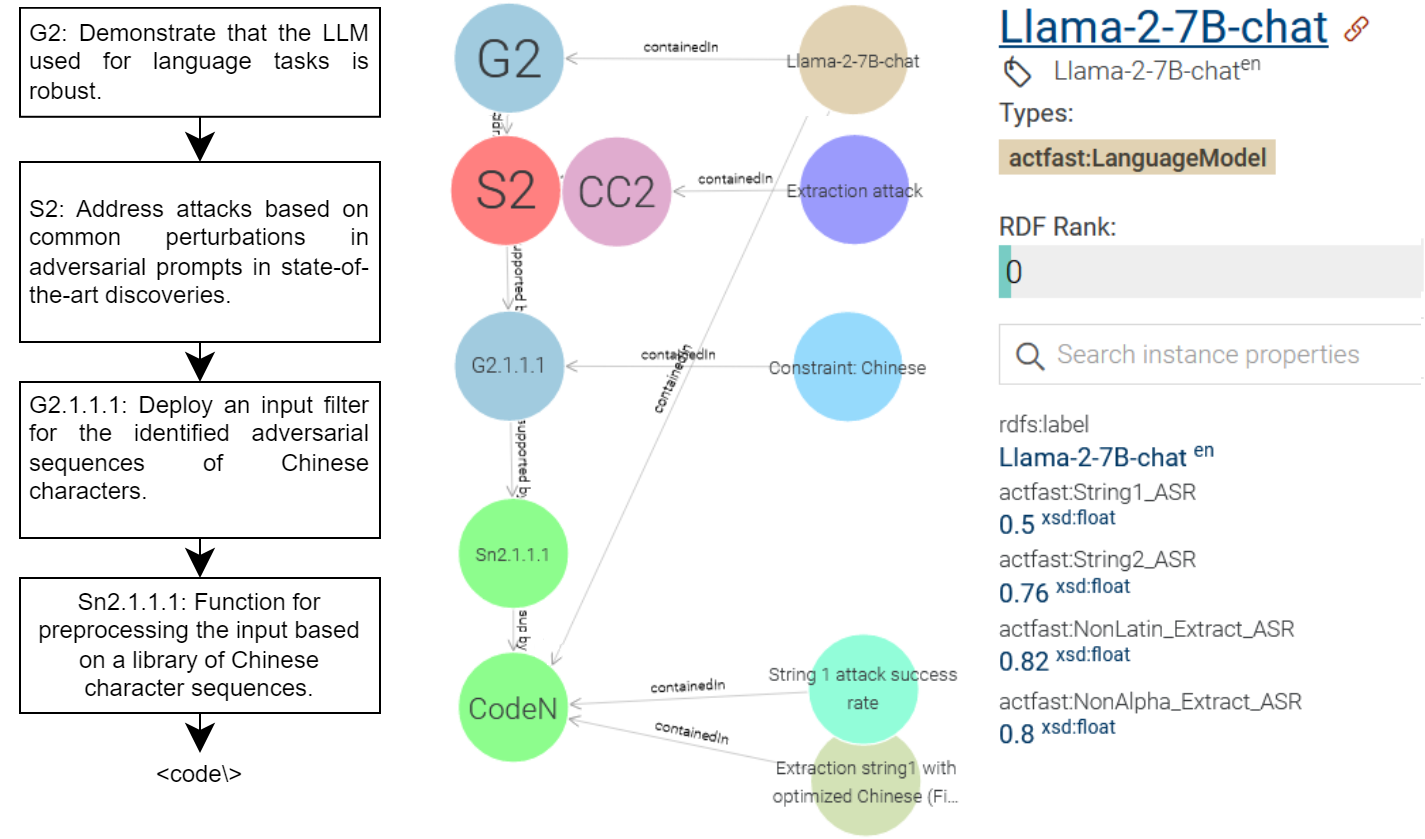}
\caption{Assurance case in GSN (left), connected graphs (center), and the queriable values in concepts (right). The boxes are a manual illustration of the text in the argument.} \label{fig2}
\end{figure}

\begin{figure}[]
\centering
\includegraphics[width=\textwidth]{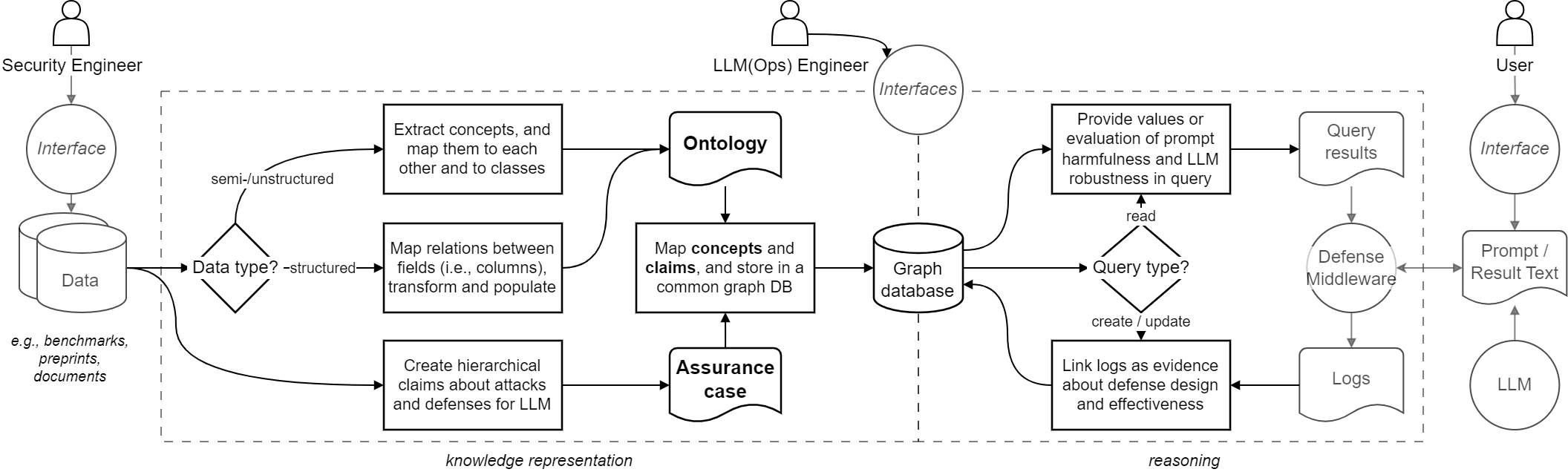}
\caption{Pipeline for representing data about attacks and defenses into an ontology and assurance case, for further reasoning by middleware implementing the LLM defenses. Original contributions emphasized.} \label{fig1}
\vspace{-15pt}
\end{figure}

The ontology allows attack- and defense-relevant values to be retrieved, calculated, and inserted with complex queries, while showing the argument and architecture to readers. We posit that this setup and pipeline (cf. Figure \ref{fig1}) separates the following maintenance concerns while providing an explainable representation of robustness: (i) explication and structuring of the approaches to defend from adversarial attacks; (ii) continuous reasoning against changes by querying the parameter values from a central repository; (iii) inserting and maintaining values in the ontology based on experiments or external empirical data; and (iv) auditing the design and effectiveness of the operationalized robustness in the LLM (cf. Figure \ref{fig2}).

\subsection{Robustness in Code Translation Tasks}

LLMs used for domain-specific language tasks can similarly be susceptible to simple adversarial attacks \cite{hajipour2023systematically}. We present a toy example where a function for calculating the factorial of a number is translated from C++ to Python. While users could attempt to jailbreak or translate intentionally harmful code, they may also be unaware of potential vulnerabilities in the input or output. These naive requests can happen with large codebases, imprecise mappings between languages, or users who lack security awareness or proficiency.

Regardless of the user's intent, the engineer could want to ensure that the LLM is not generating harmful code. Robustness would then include a sequence of specific claims (S1; G1.5) about various defenses (cf. Figure \ref{fig5}). We show claims about three example mechanisms for the given context (C1.5): perplexity input filter (G1.5.1; \cite{alon2023detecting}); and code analysis output filters to detect injections (G1.5.3; \cite{mitre2023,github2024}) or lack of input sanitization (G1.5.4; \cite{liu2023code}).

Input filters could detect malicious requests. Perplexity (Sn1.5.1) filtering is a mechanism for determining if the prompt is an outlier (e.g., a gradient-based attack) by comparing it with the properties of data on which the LLM was trained. Randomization of input may not lead to comprehensible or functioning code in the prompt, but depending on the LLM training (e.g., helpfulness, correctness) and application (e.g., code autocompletion), the LLM may still generate executable output with vulnerable, malicious or toxic elements. 

When an input filter fails to detect an attack, or the LLM generates problematic code from benign prompts, an engineer can rely on output filters. Code analysis, for example, could flag vulnerable or malicious elements with manually defined software tests \cite{mitre2023,liu2023code} or automatic queries from externally maintained tools \cite{github2024}. Such flags can be treated differently. For vulnerable code, the LLM could provide three aspects in the same output: the translated function; a warning that end-users of the function could create problems with wrong or intentionally manipulated input (CC1.5.4), unless inputs are sanitized (Sn1.5.4); and error message patterns to fix this (Sn1.5.3.1; Sn1.5.4.1). For malicious code, such as a request to translate an injection that bypassed the input filter, the filter could prevent the translation from reaching the user without affecting the helpfulness of the LLM (Sn1.5.3).

%% file: content_short/conclusion.tex
In this research-in-progress paper, we explore assuring the robustness of LLMs using human-comprehensible assurance cases and machine-comprehensible semantic networks in ontologies. We show that our approach can be implemented alongside the LLM-based system, to make its robustness explainable by providing metadata for code variables, encoding the dependencies explicitly, and making the evidence transparent. Implications for researchers include studying different types of claim and evidence, as well as notations towards a shared knowledge for LLM assurance. Implications for practitioners include a novel idea for proactively engineering adversarially robust LLMs. Future work will center on exploring and evaluating this approach with real-life implementations and industrial use cases, as well as addressing the limitation of manually formalizing arguments and ontologies, to cover various attacks and improve maintainability over time.

\begin{figure}[]
\centering
\includegraphics[width=0.9\textwidth]{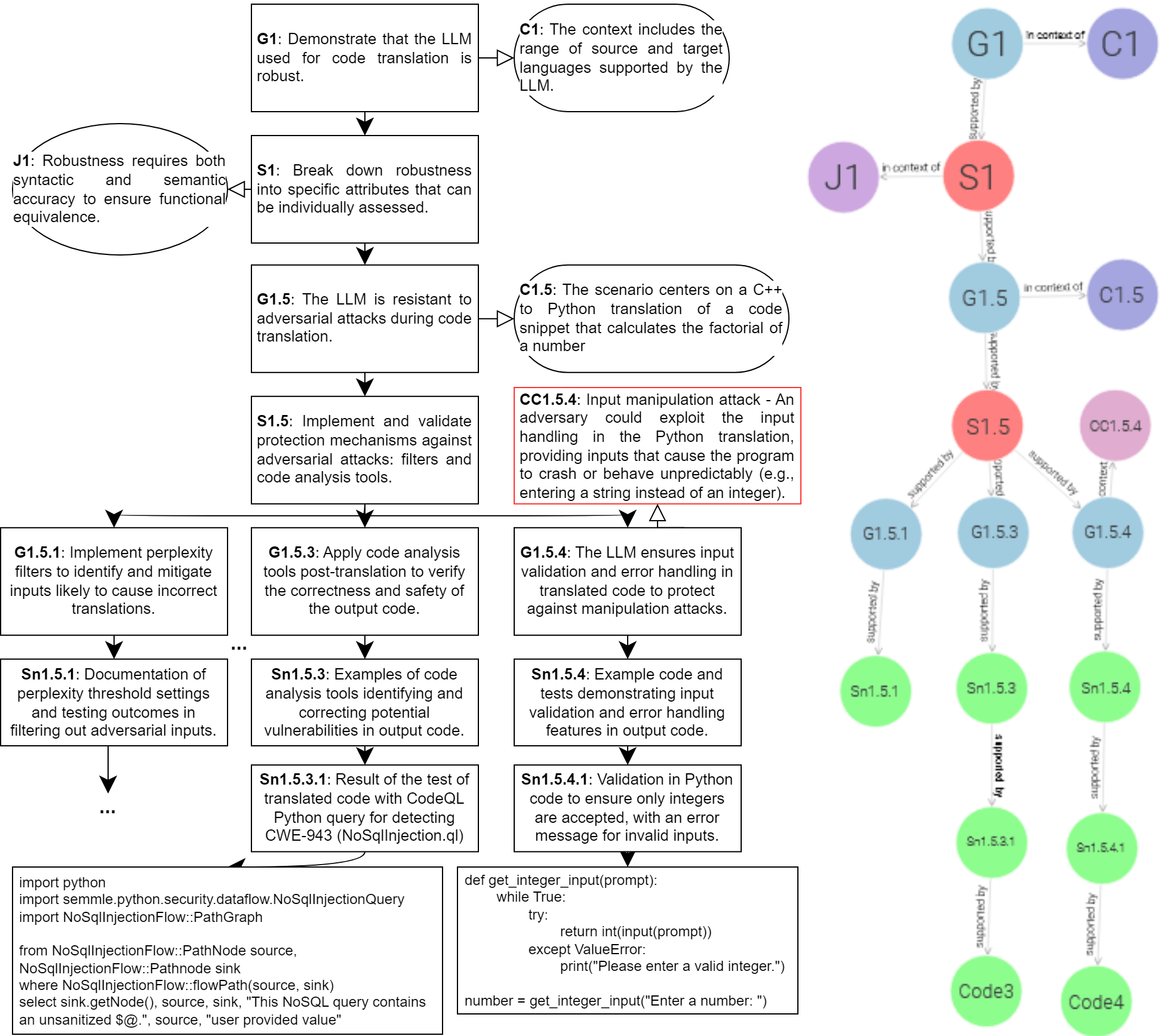}
\caption{Sketched argument (left) and its representation in ontology (right).} \label{fig5}
\end{figure}